\documentclass{article}

\PassOptionsToPackage{numbers,sort&compress}{natbib}
\usepackage[preprint]{neurips_2026}

\usepackage[utf8]{inputenc} % allow utf-8 input
\usepackage[T1]{fontenc}    % use 8-bit T1 fonts
\usepackage{hyperref}       % hyperlinks
\usepackage{url}            % simple URL typesetting
\usepackage{booktabs}       % professional-quality tables
\usepackage{amsfonts}       % blackboard math symbols
\usepackage{nicefrac}       % compact symbols for 1/2, etc.
\usepackage{microtype}      % microtypography
\usepackage{xcolor}         % colors
\usepackage{graphicx}
\usepackage{subcaption}
\usepackage{wrapfig}
\usepackage{amsmath}
\usepackage{amssymb}
\usepackage{mathtools}
\usepackage{amsthm}
\usepackage[capitalize,noabbrev]{cleveref}
\usepackage[most]{tcolorbox}
\usepackage{enumitem}
\usepackage{fontawesome5}
\usepackage{algorithm}
\usepackage{algpseudocode}

\theoremstyle{plain}

\theoremstyle{definition}

\theoremstyle{remark}

\usepackage[textsize=tiny]{todonotes}

\title{SFT-then-RL Outperforms Mixed-Policy Methods for LLM Reasoning}

\author{%
  Alexis Limozin$^{1,2}$ \quad Eduard Durech$^{1}$ \quad Torsten Hoefler$^{1}$ \\
  \textbf{Imanol Schlag}$^{1}$\thanks{Equal senior authorship.} \quad \textbf{Valentina Pyatkin}$^{1,3}$\footnotemark[1] \\[0.5em]
  $^{1}$ETH AI Center, ETH Z\"urich \quad $^{2}$EPFL \quad $^{3}$Allen Institute for AI \\[0.3em]
  Corresponding author: \texttt{alexis@limozin.net}
}

\begin{document}

\maketitle
\makeatletter
\if@anonymous\else
\vspace{-2em}
\begin{center}
\href{https://github.com/alek6kun/sft_then_rl}{\faGithub\ \texttt{alek6kun/sft\_then\_rl}}
\end{center}
\vspace{1em}
\fi
\makeatother

\begin{abstract}
Recent mixed-policy optimization methods for LLM reasoning that interleave or blend supervised and reinforcement learning signals report improvements over the standard SFT-then-RL pipeline. We show that numerous recently published research papers rely on a faulty baseline caused by two distinct bugs: a CPU-offloaded optimizer bug in DeepSpeed that silently drops intermediate micro-batches during gradient accumulation (affecting multiple downstream frameworks including TRL, OpenRLHF and Llama-Factory), and a loss aggregation bug in OpenRLHF that incorrectly weights per-mini-batch losses. Together they suppress SFT performance, with the optimizer bug accounting for most of the gap and the loss aggregation bug contributing a smaller additional effect. Once corrected, the standard SFT-then-RL pipeline surpasses every published mixed-policy method we evaluate by +3.8 points on math benchmarks with Qwen2.5-Math-7B and by +22.2 points with Llama-3.1-8B. Even a truncated variant with just 50 RL steps outperforms mixed-policy methods on math benchmarks while using fewer FLOPs.
\end{abstract}

\begin{figure}[h!]
\centering
\includegraphics[width=\textwidth]{figures/teaser.pdf}
\caption{Results on Qwen2.5-Math-7B. \textbf{Left:} Training reward vs.\ compute for SFT$\to$RL, and the buggy SFT$\to$RL. \textbf{Right:} SFT evaluation score as bugs are progressively fixed (error bars: std over 3 runs).}
\label{fig:teaser}
\end{figure}

%%%%%%%%%%%%%%%%%%%%%%%%%%%%%%%%%%%%%%%
\section{Introduction}\label{sec:baseline-intro}
%%%%%%%%%%%%%%%%%%%%%%%%%%%%%%%%%%%%%%%

The ability of large language models to perform complex mathematical and scientific reasoning has improved dramatically over the past year. OpenAI's o1 series~\cite{openai2024openaio1card} first demonstrated that reinforcement learning enables models to produce chains of thought~\cite{wei2022chain} that solve competition-level problems~\cite{openai2024openaio1card}. DeepSeek-R1~\cite{Guo_2025} and Kimi k1.5~\cite{kimiteam2025kimik15scalingreinforcement} subsequently laid out open training recipes that brought these reasoning capabilities to open-weight models, exhibiting emergent behaviors such as self-verification, backtracking, and multi-step planning, and inspiring a wave of open-weight reasoning models with increasingly strong performance. In particular, the two-stage pipeline of SFT-then-RL popularized by large-scale post-training efforts~\cite{lambert2025tulu, Guo_2025} became the standard approach for training reasoning models today: supervised fine-tuning (SFT) on expert-generated chain-of-thought demonstrations to equip the model with domain knowledge and output formatting through memorization~\cite{chu2025sft, mecklenburg2024injectingnewknowledgelarge}, followed by reinforcement learning from verifiable rewards (RLVR) to sharpen the policy toward higher-reward reasoning strategies~\cite{chu2025sft, yue2025does, zhao2025echo, matsutani2026rl}.

% Direction of recent studies toward mixed-policy optimization
Recent work on post-training for LLM reasoning has moved beyond the standard SFT followed by RL pipeline toward \emph{mixed-policy} methods that interleave or blend supervised and reinforcement learning signals during training. The motivation is intuitive: on-policy RL suffers from sparse rewards on hard problems where the model rarely generates a correct solution, and SFT alone does not develop the model's own reasoning capacity. By combining on-policy rollouts with off-policy expert demonstrations, mixed-policy methods aim to leverage the complementary strengths of both: SFT's ability to learn from demonstrations beyond the model's initial capabilities, and RL's ability to refine and sharpen existing reasoning.

A growing body of methods has emerged along this direction~\cite{yan2025learning, ma2026learning, fu2026srft, huang2026blending, lv2026unifiedviewlargelanguage, yuan2025mitigatingforgettingsupervisedreinforcement, guan2025recallextenddynamicsenhancingsmall, liu2025uft, zhang2026onpolicy, liu2025superrlreinforcementlearningsupervision, chen2025stepwiseadaptiveintegrationsupervised, wu2025templaterlstructuredtemplateguidedreinforcement}, many claiming state-of-the-art results on mathematical reasoning benchmarks. Numerous methods~\cite{yan2025learning, ma2026learning, fu2026srft, yuan2025mitigatingforgettingsupervisedreinforcement, huang2026blending, lv2026unifiedviewlargelanguage, guan2025recallextenddynamicsenhancingsmall} follow the SFT setup of \citet{yan2025learning} and report improvements over SFT and SFT$\to$RL baselines. We show that their SFT baselines are weakened by bugs in shared training frameworks (\cref{sec:openrlhf-bug}), and that a correctly implemented SFT followed by RL pipeline matches or exceeds their reported scores on average. The sequential pipeline also achieves this while using fewer FLOPs (\cref{fig:teaser}, left), as SFT ensures dense reward signal for RL from the start, whereas mixed-policy methods must learn and refine simultaneously under weak early signal.

Our contributions can be summarized as follows:
\begin{itemize}
    \item We discover that numerous recently published research papers on mixed-policy training rely on a faulty baseline: we identify and fix two SFT bugs -- a CPU-offloaded optimizer bug in DeepSpeed~\cite{DeepSpeed} that silently drops intermediate micro-batches during gradient accumulation (affecting multiple downstream training frameworks), and a loss aggregation bug in \texttt{OpenRLHF}~\cite{hu-etal-2025-openrlhf} that incorrectly weights per-mini-batch losses. Together they deflate SFT performance by up to 5.7 points with Qwen2.5-Math-7B.
    \item Once corrected, the standard SFT followed by RL pipeline achieves state-of-the-art results, surpassing the best published mixed-policy method by \textbf{+3.8} points on in-distribution tasks with Qwen2.5-Math-7B and by \textbf{+22.2} points with Llama-3.1-8B, while maintaining generalizability on out-of-distribution benchmarks. Even a truncated variant with just 50 instead of 500 RL iterations \textbf{outperforms all mixed-policy methods} on math benchmarks while using fewer FLOPs, calling their claimed improvements into question.
    \item Our findings restore confidence in the SFT followed by RL paradigm and highlight the importance of cross-framework validation, as silent bugs in widely used SFT pipelines were sufficient to systematically deflate baselines across multiple independent studies.
\end{itemize}

%%%%%%%%%%%%%%%%%%%%%%%%%%%%%%%%%%%%%%%
\section{Bugs in Widely-Used SFT Frameworks}\label{sec:openrlhf-bug}

%%%%%%%%%%%%%%%%%%%%%%%%%%%%%%%%%%%%%%%
We identify two bugs in widely used open-source training frameworks that silently degrade SFT quality. Both are triggered by distributed training configurations, making them difficult to detect without cross-framework validation.

%--------------------------------------
\subsection{CPU-Offloaded Optimizer Bug}\label{sec:optimizer-bug}
%--------------------------------------

% Incorrect gradient norm with CPU-offloaded Adam under gradient accumulation
\citet{yan2025learning} use \texttt{OpenRLHF}~\cite{hu-etal-2025-openrlhf} with DeepSpeed~\cite{DeepSpeed} ZeRO Stage~2 and CPU-offloaded Adam to reduce GPU memory consumption during SFT training. However, a bug in the CPU-offloaded gradient accumulation routine causes only the first micro-batch's gradients to reach the optimizer. Specifically, the offloading code copies gradients to CPU inside an \texttt{else} branch that executes only when \texttt{micro\_step\_id == 0} (i.e., the first micro-batch), while intermediate micro-batches accumulate gradients on GPU correctly but never trigger a copy. On the next full optimizer step, the CPU-side optimizer therefore sees only the first micro-batch's gradients rather than the accumulated sum. The fix is straightforward: moving the \texttt{copy\_gradients\_to\_cpu()} call outside the \texttt{else} branch so it executes after every micro-batch's accumulation. This bug affects any framework using DeepSpeed ZeRO Stage~1 or~2 with an offloaded optimizer, including \texttt{TRL}~\cite{vonwerra2020trl} and \texttt{Llama-Factory}~\cite{zheng2024llamafactory}; \citet{ma2026learning} use \texttt{Llama-Factory} for SFT, though they have not released their training configuration. The mixed-policy methods themselves are unaffected, as they are implemented entirely in \texttt{verl}~\cite{Sheng_2025}, which does not use DeepSpeed, the standalone SFT baselines they compare against are impacted.

This bug was introduced in September 2024 in DeepSpeed PR \#6550.\footnote{\url{https://github.com/deepspeedai/DeepSpeed/pull/6550}} We validate the fix against two independent baselines: the GPU-resident DeepSpeed optimizer in \texttt{OpenRLHF} and the PyTorch FSDP~\cite{FSDP} optimizer in \texttt{verl}. We show in \cref{sec:baseline-experiments} that the patched optimizer matches both in average loss and gradient norms. We submitted a pull request to DeepSpeed with the fix, which has been merged upstream; the patch is detailed in \cref{app:deepspeed-patch}.

%--------------------------------------
\subsection{Loss Aggregation Bug}\label{sec:loss-agg-bug}
%--------------------------------------

% Token-level vs. sequence-level loss averaging
The standard cross-entropy loss for SFT averages over all response tokens in a batch, masking out both padding and prompt tokens. As documented by~\citet{unsloth2024gradient} and~\citet{huggingface2024gradient}, a common bug in gradient accumulation computes a \emph{mean of per-mini-batch means} instead of the true per-token mean: since mini-batches contain different numbers of response tokens, this weights each mini-batch equally regardless of how many active tokens it contains. The same distortion arises across distributed data-parallel ranks, where each rank independently computes its local mean loss before averaging across ranks. \texttt{OpenRLHF}~\cite{hu-etal-2025-openrlhf}, \texttt{Llama-Factory}~\cite{zheng2024llamafactory}, and early versions of \texttt{verl}~\cite{Sheng_2025} all exhibit this bug. The fix requires aggregating token-level loss sums and counts across all ranks and mini-batches before dividing. \texttt{Verl} fixed this in November 2025\footnote{\url{https://github.com/verl-project/verl/pull/3994}}; \texttt{OpenRLHF} and \texttt{Llama-Factory} have not as of the time of writing. We have submitted a pull request to \texttt{OpenRLHF} with the fix; the pseudocode is detailed in \cref{app:openrlhf-patch}.

This bug is inherited from pretraining codebases, where data packing ensures equal active token counts across ranks, making the mean-of-means equivalent to the true per-token mean. In SFT, however, prompts and responses vary in length across samples, so mini-batches and ranks almost always have different active token counts, meaning the bug affects almost every training step. Its impact on SFT performance is smaller than the CPU-offloaded optimizer bug, but not negligible: as shown in \cref{tab:openrlhf-reproduction}, fixing the loss aggregation bug on top of the optimizer fix still yields a measurable improvement and stabilizes loss variability (see \cref{sec:baseline-experiments}).

%%%%%%%%%%%%%%%%%%%%%%%%%%%%%%%%%%%%%%%
\section{Experimental Setup}\label{sec:baseline-setup}
%%%%%%%%%%%%%%%%%%%%%%%%%%%%%%%%%%%%%%%

% Explain main reproduction methods like the papers
We reproduce the SFT baselines and mixed-policy methods under controlled conditions to isolate the impact of the bugs described in \cref{sec:openrlhf-bug}. All methods share the same base model, dataset, and evaluation protocol, which we detail below before describing the method-specific setups.

\paragraph{Training dataset.} The training dataset is OpenR1-Math-46k-8192~\cite{yan2025learning}, a length-filtered subset of OpenR1-Math-220k~\cite{openr1} whose prompts are drawn from NuminaMath~1.5~\cite{li2024numinamath} and whose off-policy reasoning traces are generated by DeepSeek-R1~\cite{Guo_2025}. \citet{yan2025learning} filter out generations longer than 8192 tokens and those verified incorrect by Math-Verify\footnote{\url{https://github.com/huggingface/Math-Verify}}. The dataset contains approximately 46k prompts paired with high-quality demonstrations. We use the full dataset for both SFT (prompts paired with demonstrations) and RL (prompts paired with ground-truth answers for reward verification). This dataset is also used by LUFFY~\cite{yan2025learning}, ReLIFT~\cite{ma2026learning}, SRFT~\cite{fu2026srft}, Prefix-RFT~\cite{huang2026blending}, and HPT~\cite{lv2026unifiedviewlargelanguage} as their common training set.

\paragraph{Models.} We follow previous works~\cite{yan2025learning, ma2026learning, fu2026srft, yuan2025mitigatingforgettingsupervisedreinforcement, huang2026blending, lv2026unifiedviewlargelanguage} and use Qwen2.5-Math-7B~\cite{yang2024qwen25mathtechnicalreportmathematical} with an increased max context length from 4096 to 16,384 and an increased RoPE theta from 10,000 to 40,000 as the base model. We also run experiments on Llama-3.1-8B~\cite{grattafiori2024llama3herdmodels} to verify generalization across model families.

\paragraph{Evaluation.} Following the evaluation protocol used by the mixed-policy methods~\cite{yan2025learning, ma2026learning, fu2026srft, huang2026blending, lv2026unifiedviewlargelanguage}, we evaluate on six mathematical reasoning benchmarks: AIME24, AIME25, AMC~\cite{li2024numinamath}, MATH-500~\cite{hendrycks2021measuring}, Minerva~\cite{lewkowycz2022solving}, and OlympiadBench (Olympiad)~\cite{he-etal-2024-olympiadbench}. For benchmarks with limited sample sizes (AIME24/25 and AMC), we report avg@32; for the rest, we use pass@1. We also evaluate on the ARC-Challenge (ARC-c)~\cite{clark2018thinksolvedquestionanswering}, GPQA-Diamond (GPQA)~\cite{rein2024gpqa}, and MMLU-Pro~\cite{wang2024mmlupro} to assess out-of-distribution generalization, as these benchmarks cover science and general knowledge domains outside the mathematical training distribution from OpenR1-Math-46k-8192. For all multiple-choice questions, we randomly shuffle the option order to mitigate information leakage. The evaluation temperature is 0.6 with a maximum response length of 8192 tokens. We use Math-Verify as the verifier during training and for testing. For each method that uses SFT, we run 3 independent training runs with different random seeds; for SFT$\to$RL pipelines, both the SFT and the subsequent RL stages use different seeds across runs, so each of the 3 runs are independent end-to-end.

\begin{table}[t]
\centering
\caption{Corrected baselines vs.\ mixed-policy methods on Qwen2.5-Math-7B (mean $\pm$ std over 3 seeds for our baselines). Left: in-distribution (ID) benchmarks. Right: out-of-distribution (OOD) benchmarks. \textbf{Bold} indicates best results, \underline{underline} indicates second-best.}
\vspace{0.5em}
\label{tab:baseline-comparison}
\resizebox{\textwidth}{!}{%
\setlength{\tabcolsep}{3pt}
\begin{tabular}{lcccccc|c||ccc|c}
\toprule
\textbf{Method} & \textbf{AIME 24} & \textbf{AIME 25} & \textbf{AMC} & \textbf{MATH-500} & \textbf{Minerva} & \textbf{Olympiad} & \textbf{Avg ID} & \textbf{ARC-c} & \textbf{GPQA} & \textbf{MMLU-Pro} & \textbf{Avg OOD} \\
\midrule
\multicolumn{12}{l}{\emph{Corrected baselines} (\texttt{verl})} \\
SFT & 33.6{\scriptsize$\pm$1.4} & \underline{28.2}{\scriptsize$\pm$0.7} & 65.4{\scriptsize$\pm$0.9} & \underline{90.0}{\scriptsize$\pm$0.9} & 40.6{\scriptsize$\pm$0.6} & 55.7{\scriptsize$\pm$0.6} & 52.2{\scriptsize$\pm$0.2} & 82.5{\scriptsize$\pm$0.6} & 24.6{\scriptsize$\pm$2.9} & 50.2{\scriptsize$\pm$0.3} & 52.4{\scriptsize$\pm$1.1} \\
SFT$\to$RL & \textbf{40.4}{\scriptsize$\pm$0.3} & \textbf{29.8}{\scriptsize$\pm$0.7} & \underline{73.2}{\scriptsize$\pm$1.2} & \textbf{92.0}{\scriptsize$\pm$0.2} & \underline{43.9}{\scriptsize$\pm$0.6} & \textbf{62.8}{\scriptsize$\pm$0.6} & \textbf{57.0}{\scriptsize$\pm$0.3} & \underline{84.0}{\scriptsize$\pm$0.7} & 40.6{\scriptsize$\pm$1.8} & \underline{55.1}{\scriptsize$\pm$0.6} & \underline{59.9}{\scriptsize$\pm$0.9} \\
\midrule
\multicolumn{12}{l}{\emph{Mixed-policy methods (reproduction attempt)}} \\
LUFFY~\cite{yan2025learning} & 25.0 & 16.9 & 64.2 & 85.2 & 35.7 & 50.8 & 46.3 & 81.4 & 41.8 & 50.9 & 58.0 \\
ReLIFT~\cite{ma2026learning} & 25.6 & 21.0 & 63.0 & 86.2 & 41.5 & 55.6 & 48.8 & 80.7 & 36.7 & 51.3 & 56.2 \\
\midrule
\multicolumn{12}{l}{\emph{Mixed-policy methods (reported)}} \\
LUFFY~\cite{yan2025learning} & 29.4 & 23.1 & 65.6 & 87.6 & 37.5 & 57.2 & 50.1 & 80.5 & 39.9 & 53.0 & 57.8 \\
ReLIFT~\cite{ma2026learning} & 28.3 & 22.9 & 65.1 & 87.9 & -- & 57.3 & -- & 81.6 & 43.1 & 53.9 & 59.5 \\
SRFT~\cite{fu2026srft} & \underline{35.3} & 21.6 & \textbf{74.3} & 89.8 & 39.7 & \underline{58.3} & \underline{53.2} & \textbf{85.3} & \textbf{46.4} & \textbf{55.9} & \textbf{62.5} \\
Prefix-RFT~\cite{huang2026blending} & 31.8 & 26.4 & 68.2 & 88.4 & 40.3 & 55.7 & 51.8 & \underline{84.0} & 39.1 & 52.1 & 58.4 \\
HPT~\cite{lv2026unifiedviewlargelanguage} & 33.0 & 21.9 & 69.4 & 89.2 & \textbf{46.0} & 56.9 & 52.7 & 81.6 & \underline{42.9} & 52.5 & 59.0 \\
\bottomrule
\end{tabular}}%
\end{table}

\paragraph{Baselines.} We compare \textbf{SFT$\to$RL}, which first fine-tunes on expert traces then runs GRPO for 500 steps, against five mixed-policy methods. All methods use Qwen2.5-Math-7B as the base model and train on the OpenR1-Math-46k-8192 dataset with 8 rollouts per prompt; every method trains for 500 steps. \textbf{LUFFY}~\cite{yan2025learning}, which mixes off-policy expert traces into on-policy GRPO; \textbf{ReLIFT}~\cite{ma2026learning}, which interleaves SFT and RL phases; \textbf{SRFT}~\cite{fu2026srft}, which jointly optimizes SFT and RL losses within each batch; \textbf{Prefix-RFT}~\cite{huang2026blending}, which samples prefixes from expert demonstrations for on-policy continuation; and \textbf{HPT}~\cite{lv2026unifiedviewlargelanguage}, which dynamically switches between SFT and GRPO based on rollout accuracy. Full descriptions and discussion of each method's baseline issues are in \cref{sec:mixed-methods}; our SFT and RL hyperparameters are in \cref{app:technical-setup}. We exclude \textbf{MIFO}~\cite{yuan2025mitigatingforgettingsupervisedreinforcement} and \textbf{RED}~\cite{guan2025recallextenddynamicsenhancingsmall}, which share the same training data but are not directly comparable: the former reports on different evaluation benchmarks, and the latter does not evaluate on our model families.

\paragraph{Reproduction of OpenRLHF vs.\ verl training.} We train four SFT variants using \texttt{OpenRLHF}, progressively fixing the loss aggregation bug, the CPU-offloaded optimizer bug, or both, and compare against an independently implemented \texttt{verl} baseline to isolate each bug's contribution on a representative subset of benchmarks (\cref{tab:openrlhf-reproduction}). For this reproduction we use a single node rather than the 4 nodes specified in our technical setup (\cref{sec:hparamssft}), since the number of nodes affects the reported gradient norm.

\paragraph{Reproduction of LUFFY and ReLIFT.} We select LUFFY and ReLIFT as the two mixed-policy methods to fully reimplement into a recent version of the \texttt{verl} codebase we use, using their original hyperparameters. Due to the high computational cost of each run, we report single-seed results; the performance gaps substantially exceed the variance observed across our 3-seed baselines. Because \texttt{verl} does not use DeepSpeed, the optimizer bug affects only the standalone SFT baselines these methods compare against, not the methods themselves, deflating the baseline side of every comparison. The same asymmetry applies to the other mixed-policy methods. We do, however, patch the loss aggregation issue (\cref{sec:loss-agg-bug}) in LUFFY's RL component and in ReLIFT's SFT and RL stage losses.

\paragraph{Chat template.}\label{sec:chat_template}
Prior works~\cite{yan2025learning, ma2026learning, fu2026srft, yuan2025mitigatingforgettingsupervisedreinforcement, huang2026blending, lv2026unifiedviewlargelanguage, guan2025recallextenddynamicsenhancingsmall} use a simplified template for Llama-3.1-8B, reporting that it cannot follow the full Qwen system prompt~\cite{yan2025learning}. With a correctly implemented SFT stage, Llama-3.1-8B follows the full prompt without issue, so we use the same template for both model families (\cref{app:chat-template}). This is consistent with the hypothesis that the inability to follow the system prompt was a symptom of undertrained SFT rather than a model limitation.

%%%%%%%%%%%%%%%%%%%%%%%%%%%%%%%%%%%%%%%
\section{Main Experiments}\label{sec:baseline-experiments}
%%%%%%%%%%%%%%%%%%%%%%%%%%%%%%%%%%%%%%%

\cref{tab:baseline-comparison} compares the corrected baselines against published mixed-policy methods. On in-distribution tasks, the corrected SFT alone (52.2) already surpasses LUFFY (46.3), ReLIFT (48.8), and Prefix-RFT (51.8); adding RL pushes the average to 57.0, outperforming the next-best method, SRFT (53.2), by \textbf{+3.8} points. The perceived gains stem from comparison against deflated SFT baselines, caused by framework bugs for LUFFY and ReLIFT (\cref{tab:openrlhf-reproduction}), and by suboptimal hyperparameters for SRFT (\cref{tab:sft-hparams}). Prefix-RFT~\cite{huang2026blending} inherits the same deflated baselines from LUFFY~\cite{yan2025learning}. On out-of-distribution tasks, SFT$\to$RL (59.9) is second only to SRFT (62.5), suggesting the pipeline generalizes well beyond math reasoning benchmarks.
\paragraph{Decomposing the baseline gap.}
\cref{tab:openrlhf-reproduction} isolates the contribution of each bug in the \texttt{OpenRLHF} SFT pipeline. Fixing only the loss aggregation bug yields a modest improvement from 48.3 to 49.1 average score, but fixing only the CPU-offloaded optimizer bug recovers nearly the entire gap, raising the average from 48.3 to 53.4. Applying both fixes together reaches 54.0: the optimizer bug accounts for the larger share of the gap, with the loss aggregation bug contributing an additional improvement on top. The fully patched \texttt{OpenRLHF} matches the 53.8 average score achieved by the independently implemented \texttt{verl} SFT.

\begin{figure}[t]
\centering
\includegraphics[width=\linewidth]{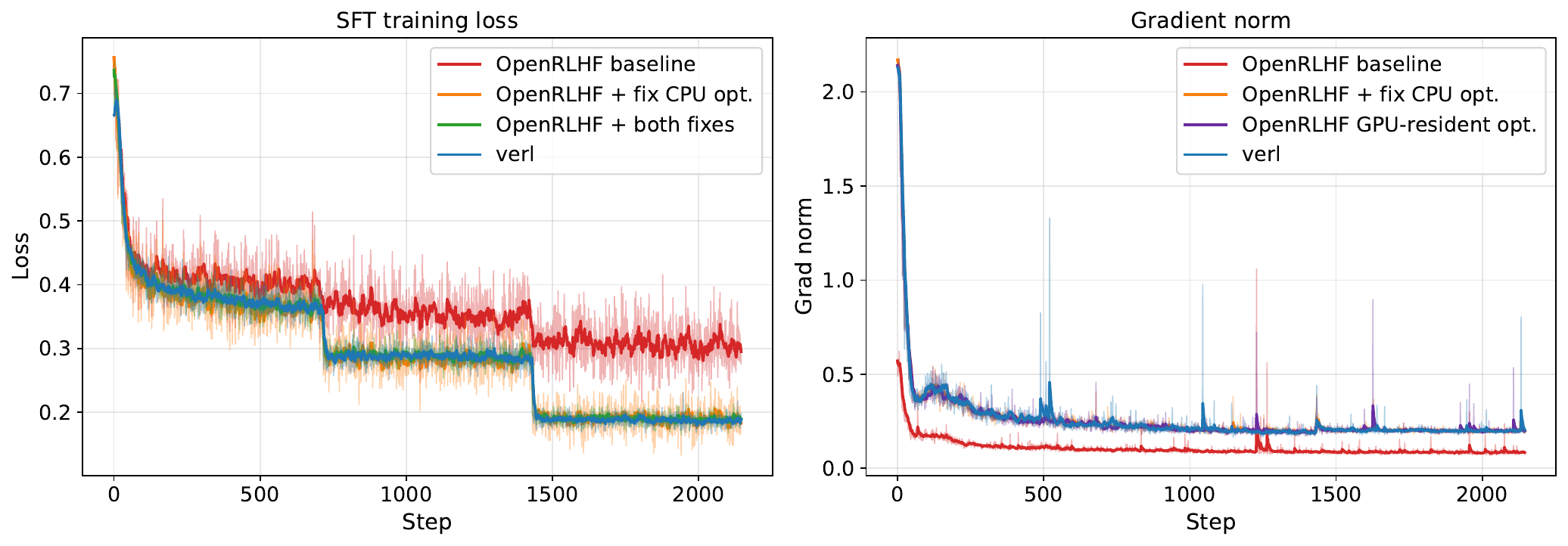}
\caption{SFT training stability across configurations. \textbf{Left:} training loss. \textbf{Right:} gradient norm. Each bug has a distinct effect on the loss: the aggregation bug introduces variability while the optimizer bug shifts the mean. Only both fixes together match the \texttt{verl} baseline. For gradient norms, only the optimizer bug has an effect, suppressing norms well below the \texttt{verl} reference, because only the first micro-batch's gradients reach the optimizer.}
\label{fig:loss-stability}
\end{figure}

\cref{fig:loss-stability} shows that, beyond final scores, the fixes also stabilize training dynamics: the buggy \texttt{OpenRLHF} SFT exhibits both a shifted mean loss and high variability; fixing the CPU-offloaded optimizer corrects the mean level but leaves the variability; fixing the loss aggregation alone reduces the variability but does not shift the mean; applying both brings the loss curve in line with \texttt{verl} (correct mean, low variability). For gradient norms, only the optimizer bug has an effect: the buggy configuration reports substantially lower gradient norms, while the patched optimizer matches both the GPU-resident \texttt{OpenRLHF} optimizer and the \texttt{verl} baseline. This stability carries over to the subsequent RL phase: RL initialized from a correctly trained SFT checkpoint starts with a higher initial reward and achieves a higher final score.

\begin{table}[t]
\centering
\small
\caption{Impact of \texttt{OpenRLHF} bugs on SFT performance on Qwen2.5-Math-7B (mean $\pm$ std over 3 seeds for our baselines). Each row isolates the contribution of each bug fix. The CPU-offloaded optimizer bug accounts for nearly all of the performance gap; fixing the loss aggregation bug alone has a smaller effect.}
\vspace{0.5em}
\label{tab:openrlhf-reproduction}
\resizebox{\textwidth}{!}{%
\begin{tabular}{lcccccc|c}
\toprule
\textbf{SFT Configuration} & \textbf{AIME 24} & \textbf{AIME 25} & \textbf{AMC} & \textbf{MATH-500} & \textbf{Olympiad} & \textbf{MMLU-Pro} & \textbf{Avg} \\
\midrule
\texttt{OpenRLHF} baseline & 25.8{\scriptsize$\pm$0.7} & 25.2{\scriptsize$\pm$0.5} & 59.0{\scriptsize$\pm$1.1} & 85.9{\scriptsize$\pm$1.0} & 50.4{\scriptsize$\pm$1.8} & 43.8{\scriptsize$\pm$1.8} & 48.3{\scriptsize$\pm$0.8} \\
+ fix loss aggregation & 28.7{\scriptsize$\pm$2.4} & 24.4{\scriptsize$\pm$0.3} & 60.1{\scriptsize$\pm$1.2} & 85.5{\scriptsize$\pm$1.0} & 51.8{\scriptsize$\pm$0.7} & 44.0{\scriptsize$\pm$1.7} & 49.1{\scriptsize$\pm$0.9} \\
+ fix optimizer & 35.1{\scriptsize$\pm$0.9} & 26.9{\scriptsize$\pm$0.3} & 64.9{\scriptsize$\pm$1.3} & 88.6{\scriptsize$\pm$0.2} & 55.3{\scriptsize$\pm$1.1} & 49.4{\scriptsize$\pm$0.5} & 53.4{\scriptsize$\pm$0.4} \\
+ fix both & 34.9{\scriptsize$\pm$0.7} & 28.2{\scriptsize$\pm$1.4} & 65.5{\scriptsize$\pm$0.6} & 89.7{\scriptsize$\pm$1.1} & 55.6{\scriptsize$\pm$0.7} & 49.8{\scriptsize$\pm$0.8} & 54.0{\scriptsize$\pm$0.2} \\
\midrule
\texttt{verl} & 33.6{\scriptsize$\pm$1.4} & 28.2{\scriptsize$\pm$0.7} & 65.4{\scriptsize$\pm$0.9} & 90.0{\scriptsize$\pm$0.9} & 55.7{\scriptsize$\pm$0.6} & 50.2{\scriptsize$\pm$0.3} & 53.8{\scriptsize$\pm$0.1} \\
\bottomrule
\end{tabular}}
\end{table}

\paragraph{Hyperparameter sensitivity.}\label{sec:hparms_sens}
Beyond framework bugs, the choice of SFT hyperparameters also affects baseline strength. Our SFT configuration (\cref{sec:hparamssft}) follows \citet{yan2025learning} and \citet{ma2026learning}, both using the same settings. \citet{fu2026srft}, however, uses a batch size of 128, a learning rate of $5 \times 10^{-6}$ with a linear schedule and 10\% warmup. We reproduced the SFT baseline with \citet{fu2026srft}'s reported hyperparameters and obtained a weaker baseline, with the average score dropping from 53.8 to 48.3 (\cref{tab:sft-hparams}). Switching to the LUFFY/ReLIFT hyperparameters recovers this gap, which in turn reduces the margin between the SFT baseline and the final result of \citet{fu2026srft} (55.9 average score on this set of evaluations; see \cref{tab:baseline-comparison}) from 7.6 points down to 2.1 points.

\begin{table}[t]
\centering
\small
\caption{Impact of SFT hyperparameters on baseline performance on Qwen2.5-Math-7B (mean $\pm$ std over 3 seeds for our baselines). Reproducing the SFT stage of \citet{fu2026srft} with their hyperparameters (10x lower learning rate, 2x batch size, linear schedule) yields a weaker baseline; switching to the LUFFY/ReLIFT hyperparameters recovers 5.5 points on average.}
\vspace{0.5em}
\label{tab:sft-hparams}
\resizebox{\textwidth}{!}{%
\begin{tabular}{lcccccc|c}
\toprule
\textbf{SFT Baseline} & \textbf{AIME 24} & \textbf{AIME 25} & \textbf{AMC} & \textbf{MATH-500} & \textbf{Olympiad} & \textbf{MMLU-Pro} & \textbf{Avg} \\
\midrule
\citet{fu2026srft} reported & 31.1 & 20.3 & 62.8 & 85.2 & 53.3 & 45.7 & 49.7 \\
\citet{fu2026srft} reproduced & 26.0{\scriptsize$\pm$0.1} & 23.5{\scriptsize$\pm$1.1} & 58.7{\scriptsize$\pm$0.4} & 85.5{\scriptsize$\pm$1.1} & 50.0{\scriptsize$\pm$0.6} & 46.5{\scriptsize$\pm$0.3} & 48.3{\scriptsize$\pm$0.2} \\
Tuned (ours) & 33.6{\scriptsize$\pm$1.4} & 28.2{\scriptsize$\pm$0.7} & 65.4{\scriptsize$\pm$0.9} & 90.0{\scriptsize$\pm$0.9} & 55.7{\scriptsize$\pm$0.6} & 50.2{\scriptsize$\pm$0.3} & 53.8{\scriptsize$\pm$0.1} \\
\bottomrule
\end{tabular}}
\end{table}

\subsection{Extension to Llama-3.1-8B}\label{sec:llama-results}

To verify that our findings are not specific to Qwen2.5-Math-7B, we repeat the experiment on Llama-3.1-8B. \cref{tab:llama-comparison} reports results on the benchmark subset for which all methods provide scores. The pattern is even more pronounced: the corrected SFT$\to$RL baseline achieves 43.7 average score, outperforming every published mixed-policy method by \textbf{+22.2} points, with the closest competitor being HPT at 21.5. The corrected SFT baseline alone reaches 33.9, already exceeding all mixed-policy results by \textbf{+12.4} points.

The failure of mixed-policy methods on Llama is particularly instructive. Unlike Qwen, which is pre-trained with emphasis on math data, Llama is a general-purpose model with little mathematical reasoning in its pre-training distribution~\cite{shao2026spuriousrewardsrethinkingtraining}, making the bootstrapping role of SFT far more critical. The training dynamics in \cref{fig:training-dynamics} (bottom row) make this concrete: with SFT$\to$RL, the RL phase starts at approximately 60\% training reward because the preceding SFT stage has already injected the mathematical knowledge that Llama lacks natively. Mixed-policy methods, by contrast, start near zero reward and improve only slowly: LUFFY's training reward remains largely flat over 500 steps, indicating that the off-policy demonstrations provide a learning signal but one far too sparse to match the dense reward that SFT$\to$RL enjoys from the start.

This points to a fundamental efficiency gap: mixed-policy methods must simultaneously bootstrap mathematical reasoning and refine it through RL, but the RL signal is extremely weak when the model cannot yet produce correct solutions on its own. The off-policy demonstrations provide some supervision, but it is diluted by on-policy rollouts that carry little to no reward signal early in training. SFT-then-RL cleanly separates these two objectives: SFT first bootstraps the model into a regime where it reliably generates correct solutions, and RL then refines this already-capable policy with dense reward signal from the start. While mixed-policy methods may eventually converge to comparable scores given enough steps, they do so far less efficiently because the RL component contributes minimally until the model has acquired sufficient knowledge -- precisely the knowledge that a dedicated SFT stage provides upfront.

\begin{table}[t]
\centering
\small
\caption{Corrected baselines versus published mixed-policy methods on Llama-3.1-8B. We report the subset of benchmarks for which all methods provide a score. \textbf{Bold} indicates best results, \underline{underline} indicates second-best. * denotes results taken from the respective papers, which use a different chat template (see \cref{sec:chat_template}).}
\vspace{0.5em}
\label{tab:llama-comparison}
\begin{tabular}{lccccc|c}
\toprule
\textbf{Method} & \textbf{AIME 24} & \textbf{AMC} & \textbf{MATH-500} & \textbf{Minerva} & \textbf{Olympiad} & \textbf{Avg} \\
\midrule
SFT (ours) & \underline{7.0}{\scriptsize$\pm$1.1} & \underline{40.4}{\scriptsize$\pm$0.6} & \underline{68.4}{\scriptsize$\pm$2.3} & 17.3{\scriptsize$\pm$0.9} & \underline{36.2}{\scriptsize$\pm$1.1} & \underline{33.9}{\scriptsize$\pm$0.9} \\
SFT$\to$RL (ours) & \textbf{16.5}{\scriptsize$\pm$0.8} & \textbf{53.9}{\scriptsize$\pm$1.3} & \textbf{78.6}{\scriptsize$\pm$1.2} & \textbf{20.7}{\scriptsize$\pm$1.5} & \textbf{48.5}{\scriptsize$\pm$0.1} & \textbf{43.7}{\scriptsize$\pm$0.2} \\
\midrule
LUFFY~\cite{yan2025learning} & 0.8 & 13.1 & 34.6 & 12.5 & 10.8 & 14.4 \\
ReLIFT~\cite{ma2026learning} & 0.6 & 14.5 & 35.4 & 14.7 & 12.8 & 15.6 \\
SRFT*~\cite{fu2026srft} & 1.9 & 14.3 & 40.1 & 15.3 & 9.5 & 16.2 \\
Prefix-RFT*~\cite{huang2026blending} & 1.3 & 13.3 & 40.6 & 18.1 & 11.9 & 17.0 \\
HPT*~\cite{lv2026unifiedviewlargelanguage} & 2.1 & 18.6 & 47.8 & \underline{18.8} & 20.4 & 21.5 \\
\bottomrule
\end{tabular}
\end{table}

\subsection{Training Dynamics}\label{sec:training-dynamics}
\begin{figure}[t]
    \centering
    \includegraphics[width=\linewidth]{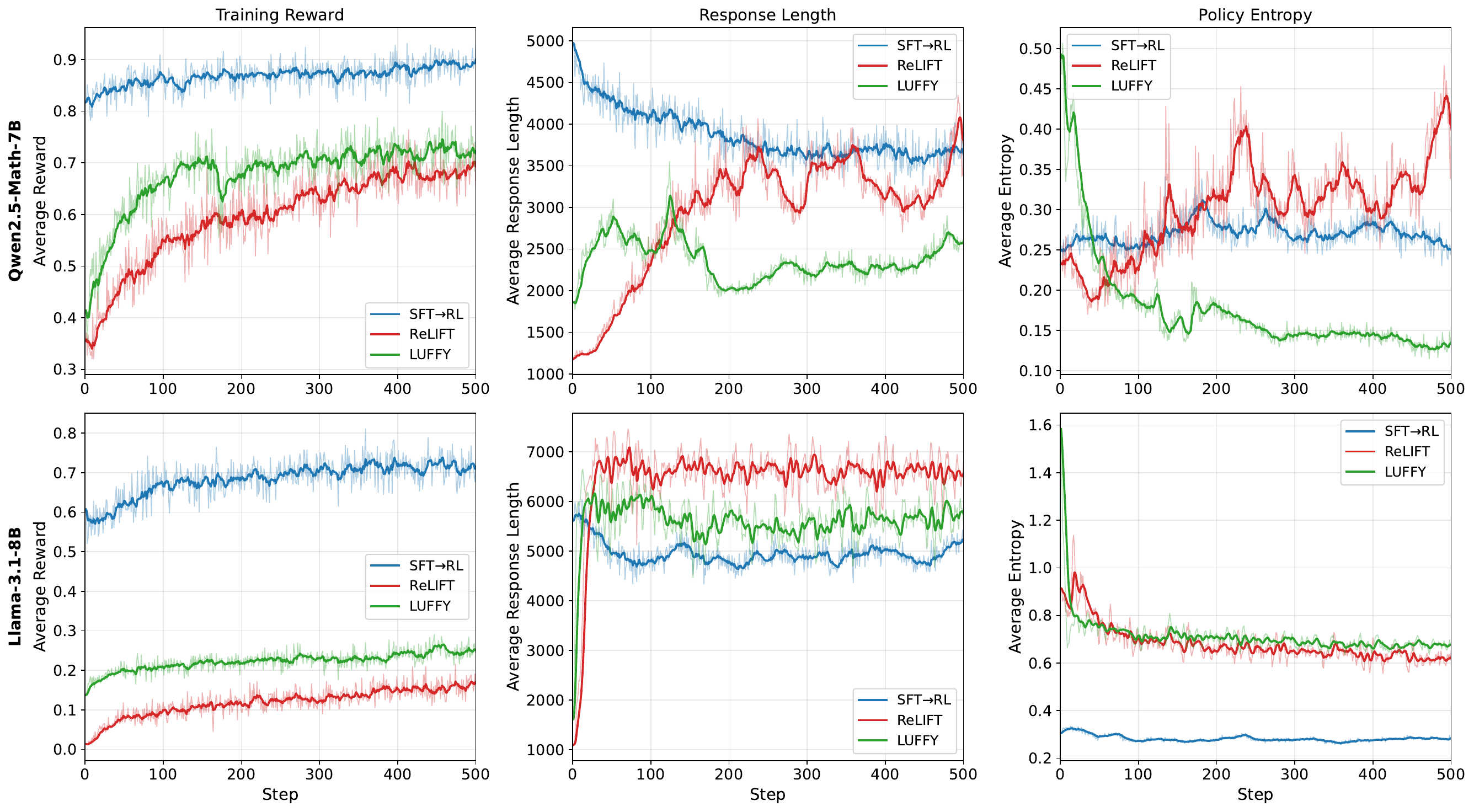}
    \caption{Training dynamics comparison between the RL part of SFT$\to$RL, LUFFY, and ReLIFT. Top row: Qwen2.5-Math-7B. Bottom row: Llama-3.1-8B. Left to right: training reward, mean response length, and policy entropy.}
    \label{fig:training-dynamics}
\end{figure}

\cref{fig:training-dynamics} compares the training dynamics of the RL phase of SFT$\to$RL against ReLIFT and LUFFY on both models. For Qwen, the RL phase of SFT$\to$RL starts above 80\% training reward, because the preceding SFT stage has already internalized expert knowledge, and continues to improve. Mixed-policy methods~\cite{yan2025learning, ma2026learning} never reach the reward level that SFT$\to$RL begins with, suggesting that interleaving off- and on-policy signals within a single stage is less efficient than performing them sequentially, or even than performing SFT alone. On Llama-3.1-8B, the gap is starker: LUFFY and ReLIFT never exceed 30\% training reward after 500 steps, while SFT$\to$RL starts at approximately 60\% and continues climbing.

For response length, the RL phase of SFT$\to$RL starts at approximately 5k tokens on Qwen, reflecting the verbose style inherited from SFT on DeepSeek-R1 demonstrations, and gradually converges toward 3.7k tokens as RL prunes unnecessary verbosity while retaining problem-solving capability. Mixed-policy methods instead show gradually increasing lengths, which \citet{ma2026learning} interpret as developing more thorough reasoning. On Llama, mixed-policy response lengths quickly diverge toward the context limit.

For entropy, \citet{yan2025learning} and \citet{ma2026learning} emphasize that their approaches maintain higher entropy compared to pure RL, which they attribute to sustained exploration. In SFT$\to$RL, the entropy remains relatively constant for both models, consistent with RL refining an already strong policy rather than exploring from scratch.

%--------------------------------------
\subsection{Training Efficiency}\label{sec:training-efficiency}
%--------------------------------------

Since the post-SFT Qwen model already exhibits strong performance, we investigate whether the RL phase can be shortened without significant degradation.
Concretely, we increase the learning rate from $1 \times 10^{-6}$ to $5 \times 10^{-6}$ and train for only 50 RL steps: 10$\times$ fewer than the default setup described in \cref{sec:hparamsrl}.

\cref{tab:fast_rl_id} summarizes the results.
On in-distribution benchmarks, the shortened schedule achieves an average score of 55.6, only 1.4 points behind the full 500-step run and still \textbf{+2.4} points ahead of the next-best mixed-policy baseline (SRFT at 53.2, see \cref{tab:baseline-comparison}).
On out-of-distribution benchmarks, the gap between the full run (59.9) and the short run (59.2) is similarly narrow at 0.7 points, while remaining competitive with mixed-policy baselines (only surpassed by SRFT at 62.5 and ReLIFT at 59.5).
These results suggest that, given a strong SFT initialization, the majority of RL gains can materialize within the early stages of RL training.

\begin{table}[t]
\centering
\caption{Faster RL on Qwen2.5-Math-7B. Left: ID benchmarks. Right: OOD benchmarks.}
\vspace{0.5em}
\label{tab:fast_rl_id}
\resizebox{\textwidth}{!}{%
\setlength{\tabcolsep}{3pt}
\begin{tabular}{lcccccc|c||ccc|c}
\toprule
\textbf{Method} & \textbf{AIME 24} & \textbf{AIME 25} & \textbf{AMC} & \textbf{MATH-500} & \textbf{Minerva} & \textbf{Olympiad} & \textbf{Avg ID} & \textbf{ARC-c} & \textbf{GPQA} & \textbf{MMLU-Pro} & \textbf{Avg OOD} \\
\midrule
SFT & 33.6{\scriptsize$\pm$1.4} & \underline{28.2}{\scriptsize$\pm$0.7} & 65.4{\scriptsize$\pm$0.9} & \underline{90.0}{\scriptsize$\pm$0.9} & 40.6{\scriptsize$\pm$0.6} & 55.7{\scriptsize$\pm$0.6} & 52.2{\scriptsize$\pm$0.2} & 82.5{\scriptsize$\pm$0.6} & 24.6{\scriptsize$\pm$2.9} & 50.2{\scriptsize$\pm$0.3} & 52.4{\scriptsize$\pm$1.1} \\
SFT$\to$RL (50 steps) & 37.1{\scriptsize$\pm$2.7} & 29.4{\scriptsize$\pm$0.6} & 70.3{\scriptsize$\pm$1.5} & 91.7{\scriptsize$\pm$0.8} & 44.1{\scriptsize$\pm$0.0} & 61.3{\scriptsize$\pm$0.9} & 55.6{\scriptsize$\pm$0.5} & 84.0{\scriptsize$\pm$0.5} & 39.6{\scriptsize$\pm$1.0} & 53.9{\scriptsize$\pm$0.4} & 59.2{\scriptsize$\pm$0.2} \\
SFT$\to$RL & 40.4{\scriptsize$\pm$0.3} & 29.8{\scriptsize$\pm$0.7} & 73.2{\scriptsize$\pm$1.2} & 92.0{\scriptsize$\pm$0.2} & 43.9{\scriptsize$\pm$0.6} & 62.8{\scriptsize$\pm$0.6} & 57.0{\scriptsize$\pm$0.3} & 84.0{\scriptsize$\pm$0.7} & 40.6{\scriptsize$\pm$1.8} & 55.1{\scriptsize$\pm$0.6} & 59.9{\scriptsize$\pm$0.9} \\
\bottomrule
\end{tabular}}%
\end{table}

\begin{wraptable}{r}{0.4\textwidth}
\centering
\small
\caption{Estimated training FLOPs for each method on Qwen2.5-Math-7B.}
\label{tab:flops}
\begin{tabular}{lc}
\toprule
\textbf{Method} & \textbf{FLOPs ($\times 10^{19}$)} \\
\midrule
SFT$\to$RL (50 steps) & 3.63 \\
LUFFY & 6.65 \\
ReLIFT & 8.76 \\
\bottomrule
\end{tabular}
\end{wraptable}

\paragraph{FLOPs.}
\cref{tab:flops} summarizes the results (calculation details in \cref{app:flops-details}). Our truncated SFT$\to$RL pipeline requires $3.63 \times 10^{19}$ FLOPs, fewer than LUFFY ($6.65 \times 10^{19}$) and ReLIFT ($8.76 \times 10^{19}$). SFT is cheap relative to RL as it processes tokens without the overhead of multiple rollouts, and the strong initialization it provides allows truncating RL to just 50 steps. Since SFT$\to$RL (50 steps) also outperforms both LUFFY and ReLIFT on ID benchmarks and matches OOD benchmarks (\cref{tab:baseline-comparison}), the pipeline delivers stronger results at lesser compute.

%%%%%%%%%%%%%%%%%%%%%%%%%%%%%%%%%%%%%%%
\section{Related Work}\label{sec:related-work}
%%%%%%%%%%%%%%%%%%%%%%%%%%%%%%%%%%%%%%%

%--------------------------------------
We distinguish between methods we directly evaluate in our controlled setting and those we discuss qualitatively due to differences in training setups, datasets, or evaluation protocols.\label{sec:mixed-methods}
%--------------------------------------

\textbf{LUFFY}~\cite{yan2025learning} extends GRPO by mixing on-policy rollouts with off-policy expert traces. It applies regularized importance sampling with a shaped policy $f(\pi)=\pi/(\pi+\lambda)$ to control distribution mismatch and avoid imitation collapse. LUFFY uses \texttt{OpenRLHF}~\cite{hu-etal-2025-openrlhf} for SFT, which is affected by the bugs described in \cref{sec:openrlhf-bug}. \textbf{ReLIFT}~\cite{ma2026learning} alternates between SFT and RL phases. It identifies unsolved problems via pass rate, applies SFT on expert demonstrations to inject missing knowledge, and resumes RL. This iterative process shrinks the set of hard problems over time. ReLIFT uses \texttt{Llama-Factory}~\cite{zheng2024llamafactory} for SFT, which is affected by the bugs described in \cref{sec:openrlhf-bug}. \textbf{SRFT}~\cite{fu2026srft} jointly optimizes supervised and reinforcement losses within each batch, using an adaptive entropy-based weight to balance the two signals. However, its SFT baseline uses a tenfold lower learning rate than other methods, producing a weaker starting point that inflates the apparent mixed-policy gain (see \cref{sec:hparms_sens}). \textbf{Prefix-RFT}~\cite{huang2026blending} integrates demonstrations into RFT by sampling a prefix from expert solutions and generating on-policy continuations, forming hybrid trajectories for policy updates. It adopts its SFT baselines from LUFFY, which are affected by the bugs described in \cref{sec:openrlhf-bug}. \textbf{HPT}~\cite{lv2026unifiedviewlargelanguage} unifies SFT and RL under a common policy gradient objective and uses a binary gate: SFT when all rollouts fail, GRPO otherwise. Notably, its own ablations show that pure SFT matches or beats off-policy RL for the offline signal. HPT does not detail its SFT setup, stating only that they "endeavored to follow previous works as closely as possible".

%--------------------------------------
\paragraph{Other mixed-policy methods that may warrant SFT auditing.} \textbf{UFT}~\cite{liu2025uft} and \textbf{CHORD}~\cite{zhang2026onpolicy} combine SFT and RL into a single loss, with CHORD using dynamic weighting and token-level uncertainty scaling. \textbf{SuperRL}~\cite{liu2025superrlreinforcementlearningsupervision} instead applies RL only when correct rollouts exist and falls back to SFT otherwise, avoiding signal mixing and highlighting risks of off-policy collapse.
\textbf{SASR}~\cite{chen2025stepwiseadaptiveintegrationsupervised} adaptively balances SFT and RL using a KL-based switching signal.
\textbf{RED}~\cite{guan2025recallextenddynamicsenhancingsmall} dynamically balances SFT and RL by regulating their contributions based on entropy changes and sample accuracy. It introduces an accuracy-aware policy shift and entropy-based weighting to control when the model imitates offline data versus explores with RL, aiming to stabilize training under distribution mismatch and limited exploration. \textbf{TemplateRL}~\cite{wu2025templaterlstructuredtemplateguidedreinforcement} increases exploration diversity by augmenting prompts with MCTS-derived templates, achieving large gains while remaining fully on-policy. \textbf{MIFO}~\cite{yuan2025mitigatingforgettingsupervisedreinforcement} interleaves RL and SFT by buffering hard questions for targeted SFT, while freezing RL-critical parameters during SFT phases to mitigate catastrophic forgetting.

%%%%%%%%%%%%%%%%%%%%%%%%%%%%%%%%%%%%%%%
\section{Conclusion and Limitations}\label{sec:baseline-conclusion}
%%%%%%%%%%%%%%%%%%%%%%%%%%%%%%%%%%%%%%%

We have shown that the reported gains of mixed-policy optimization methods for LLM reasoning trace back to deflated baselines rather than methodological innovation. A CPU-offloaded optimizer bug in DeepSpeed silently drops intermediate micro-batches during gradient accumulation, affecting downstream frameworks including \texttt{TRL}, \texttt{OpenRLHF} and \texttt{Llama-Factory}; fixing it alone recovers most of the gap. A second loss aggregation bug in \texttt{OpenRLHF} incorrectly weights per-mini-batch losses and, on top of adding training instability, contributes a further measurable degradation. Together the two fixes close the gap to the independently implemented \texttt{verl} baseline.

Once corrected, the standard SFT followed by RL pipeline, to the best of our knowledge, exceeds all five mixed-policy methods we evaluate on in-distribution tasks and achieves comparable or superior out-of-distribution performance, on both Qwen2.5-Math-7B and Llama-3.1-8B. The corrected pipeline is also more efficient, converging in 50 RL steps using fewer FLOPs than LUFFY or ReLIFT, because SFT bootstraps the model into a regime with dense RL reward signal rather than requiring simultaneous knowledge acquisition and refinement. These findings carry two implications: first, framework-level diversity is essential for robustness, as silent bugs in widely used SFT pipelines were sufficient to deflate baselines across multiple papers; second, empirical ML results should be cross-validated across independently implemented frameworks to guard against systematic baseline deflation. Because these bugs affect general-purpose SFT pipelines rather than mixed-policy code specifically, they could potentially invalidate research findings in other areas that rely on the same affected frameworks.

\paragraph{Limitations.} Our study focuses on mathematical reasoning with Qwen2.5-Math-7B and Llama-3.1-8B; we do not train on other domains (e.g., code, general reasoning) or larger scales, where SFT quality may matter differently. We also do not consider other model families, since the mixed-policy papers themselves do not. There is a nuance in our comparison: these methods position themselves as single-stage alternatives to SFT-then-RL, and that framing is their core contribution. Applying mixed-policy training atop a correctly trained SFT checkpoint might yield further gains, but none of these works report such a setup, as it would counter their thesis. Our results therefore invalidate the published comparisons without precluding that a mixed-policy stage could add value on top of a proper SFT baseline. Finally, for ReLIFT and HPT we could not fully verify the SFT setup: ReLIFT uses Llama-Factory but has not released its exact configuration, and HPT does not detail its SFT hyperparameters; our assessment assumes both inherit the same affected pipeline.

\newpage
\begin{ack}
We thank Nathan Ranchin, Lorenzo Paleari, and Yixuan Xu for helpful discussions. This work was conducted during Alexis Limozin’s Master thesis at the ETH AI Center, while he was a Master student at EPFL. This work was supported as part of the Swiss AI Initiative by compute grant infra01 from the Swiss National Supercomputing Centre (CSCS) on Alps.
\end{ack}
\bibliography{neurips_2026}
\bibliographystyle{unsrtnat}

%%%%%%%%%%%%%%%%%%%%%%%%%%%%%%%%%%%%%%%%%%%%%%%%%%%%%%%%%%%%%%%%%%%%%%%%%%%%%%%
%%%%%%%%%%%%%%%%%%%%%%%%%%%%%%%%%%%%%%%%%%%%%%%%%%%%%%%%%%%%%%%%%%%%%%%%%%%%%%%
% APPENDIX
%%%%%%%%%%%%%%%%%%%%%%%%%%%%%%%%%%%%%%%%%%%%%%%%%%%%%%%%%%%%%%%%%%%%%%%%%%%%%%%
%%%%%%%%%%%%%%%%%%%%%%%%%%%%%%%%%%%%%%%%%%%%%%%%%%%%%%%%%%%%%%%%%%%%%%%%%%%%%%%
\newpage
\appendix
\crefalias{section}{appendix}
\section{Technical Setup and Hyperparameters}\label{app:technical-setup}

\paragraph{Technical setup.} All experiments were conducted on 4 nodes of 4 NVIDIA GH200 GPUs (16 GPUs total). All training runs use the \texttt{verl}~\cite{Sheng_2025} framework, except for the SFT bug reproduction runs in \cref{tab:openrlhf-reproduction} which use \texttt{OpenRLHF}~\cite{hu-etal-2025-openrlhf} paired with DeepSpeed~\cite{DeepSpeed}. We use PyTorch Fully Sharded Data Parallel~\cite{FSDP} for distributed training, and \texttt{vLLM}~\cite{vllm2023} for rollout generation. The bugs described in \cref{sec:openrlhf-bug} were identified in DeepSpeed v0.18.9 and OpenRLHF v0.9.10.

\paragraph{SFT hyperparameters.}\label{sec:hparamssft}
Batch size 64, learning rate $5 \times 10^{-5}$ with cosine scheduling (10\% warmup, minimum ratio 0.1), AdamW with $\beta_1 = 0.9$, $\beta_2 = 0.999$, and weight decay 0.01, trained for 3 epochs.

\paragraph{RL hyperparameters.}\label{sec:hparamsrl}
Rollout batch size 128, 8 rollouts per prompt, mini-batch size 64, constant learning rate $1 \times 10^{-6}$, entropy coefficient 0, temperature 1.0, maximum response length 8192 tokens, trained for 500 steps. Following recent works~\cite{yu2025dapo, liu2025understanding, yan2025learning}, we use an improved version of GRPO with asymmetric clipping~\cite{yu2025dapo} ($\epsilon_{\text{low}}=0.2$ and $\epsilon_{\text{high}}=0.28$) and token-level loss~\cite{yu2025dapo}, remove the KL loss term~\cite{yu2025dapo, yan2025learning}, and remove length normalization and standard error normalization~\cite{liu2025understanding}.

\section{FLOPs Calculation Details}\label{app:flops-details}
 
To enable a fair comparison of overall training costs across methods employing different framework-level optimization strategies, we estimate FLOPs following~\cite{zhang2025bread, snell2025scaling,hoffmann2022an,sardana2024beyond}.
A forward pass through a model with $N$ parameters on $D$ tokens costs approximately $2ND$ FLOPs, and a backward pass costs roughly $4ND$; supervised finetuning therefore requires $6ND$ FLOPs per sample.
For RL, we additionally account for rollout generation by the inference engine separately from the training forward pass: each on-policy sequence is generated once by the inference engine ($2ND$ FLOPs) and then re-processed by the training engine for the policy gradient update (one forward + one backward, $6ND$ FLOPs), for a total of $8ND$ FLOPs per on-policy rollout. Off-policy expert traces, when used, incur only the training cost ($6ND$) since no generation is required.
For all methods we use $N{=}7\times10^{9}$ (Qwen2.5-Math-7B) and an average off-policy demonstration data response length of $D_{\text{data}}{=}4{,}200$ tokens.
 
\begin{itemize}[leftmargin=*]
    \item \textbf{LUFFY} trains for 500 steps at batch size 128 with 7 on-policy rollouts ($D_{\text{rollout}}{=}2{,}200$ tokens average response length) and 1 off-policy expert trace per sample question. The on-policy rollouts cost $8 \times 7 \times N \times D_{\text{rollout}} = 8.62 \times 10^{14}$ (generation + training), and the expert trace costs $6 \times N \times D_{\text{data}} = 1.76 \times 10^{14}$ (training only), totalling $1.04 \times 10^{15}$ FLOPs per sample and $\mathbf{6.65 \times 10^{19}}$ FLOPs overall.
    \item \textbf{ReLIFT} trains for 500 steps at batch size 128 with 8 on-policy rollouts ($D_{\text{rollout}}{=}3{,}000$ tokens average response length), costing $8 \times 8 \times N \times D_{\text{rollout}} = 1.34 \times 10^{15}$ FLOPs per sample. It also interleaves 138 SFT updates at batch size 64 ($1.56 \times 10^{18}$ FLOPs), bringing its total to $\mathbf{8.76 \times 10^{19}}$ FLOPs.
    \item \textbf{SFT\,$\boldsymbol{\to}$\,RL} finetunes for 3 epochs over $46$k samples at sequence length $D_{\text{data}}$ ($2.43 \times 10^{19}$ FLOPs), then runs 50 RL steps at batch size 128 with 8 on-policy rollouts (average response length $D_{\text{rollout}}{=}4{,}200$ tokens, contributing $8 \times 8 \times N \times D_{\text{rollout}} = 1.88 \times 10^{15}$ FLOPs per sample, $1.20 \times 10^{19}$ FLOPs total), for a combined cost of $\mathbf{3.63 \times 10^{19}}$ FLOPs.
\end{itemize}
 
\noindent All estimates count only response tokens; including prompt tokens would scale all values proportionally without affecting relative rankings.

\newpage
\section{DeepSpeed Bug Fix Patch}\label{app:deepspeed-patch}
\begin{figure}[h!]
\begin{tcolorbox}[title=Patch, colback=gray!5, colframe=gray!60, boxsep=2pt, left=4pt, right=4pt, top=2pt, bottom=2pt]
\begin{small}
\begin{verbatim}
@@ -1394,8 +1394,7 @@ class DeepSpeedZeroOptimizer(ZeROOptimizer):

         if self.micro_step_id > 0:
             accumulate_gradients()
-        else:
-            copy_gradients_to_cpu()
+        copy_gradients_to_cpu()

     def set_norm_for_param_grad(self, param):
         param_id = self.get_param_id(param)
\end{verbatim}
\end{small}
\end{tcolorbox}
\caption{Patch for \texttt{deepspeed/runtime/zero/stage\_1\_and\_2.py}}
\end{figure}

\makeatletter
\if@anonymous\else
\noindent This fix has been submitted and merged upstream: \href{https://github.com/deepspeedai/DeepSpeed/pull/7967}{\faGithub\ \texttt{deepspeedai/DeepSpeed\#7967}}.
\fi
\makeatother

\section{OpenRLHF Loss Aggregation Bug Fix Pseudocode}\label{app:openrlhf-patch}

\begin{algorithm}[h!]
\caption{Original OpenRLHF SFT loss computation (buggy).}
\label{alg:openrlhf-loss-before}
\begin{algorithmic}[1]
\Require per-token log-probs $\ell_i$, loss mask $m_i$ on local DP rank $k$
\State $\mathcal{L}_k \gets \textsc{MaskedMean}(-\ell_i,\; m_i)$ \Comment{Local mean over tokens on rank $k$}
\State \textbf{backward}$(\mathcal{L}_k)$ \Comment{Distributed optimizer averages gradients across DP ranks}
\end{algorithmic}
\end{algorithm}

\begin{algorithm}[h!]
\caption{Fixed OpenRLHF SFT loss computation.}
\label{alg:openrlhf-loss-after}
\begin{algorithmic}[1]
\Require per-token log-probs $\ell_i$, loss mask $m_i$ on local DP rank $k$, DP size $D$
\State $S_k \gets \textsc{MaskedSum}(-\ell_i,\; m_i)$ \Comment{Local loss sum}
\State $n_k \gets \sum_i m_i$ \Comment{Local token count}
\State $N \gets \textsc{AllReduce}(n_k,\; \textsc{Sum})$ \Comment{Global token count across all DP ranks}
\State $\mathcal{L}_k \gets \dfrac{S_k}{N} \times D$ \Comment{Scale by $D$ to account for gradient averaging across DP ranks}
\State \textbf{backward}$(\mathcal{L}_k)$
\end{algorithmic}
\end{algorithm}

\makeatletter
\if@anonymous\else
\noindent This fix has been submitted upstream: \href{https://github.com/OpenRLHF/OpenRLHF/pull/1216}{\faGithub\ \texttt{OpenRLHF/OpenRLHF\#1216}}.
\fi
\makeatother

\section{Chat Template}\label{app:chat-template}
\begin{figure}[h!]
\begin{tcolorbox}[title=Chat template, colback=gray!5, colframe=gray!60, boxsep=2pt, left=4pt, right=4pt, top=2pt, bottom=2pt]
\begin{small}
\textbf{System:} Your task is to follow a systematic, thorough reasoning process before providing the final solution. This involves analyzing, summarizing, exploring, reassessing, and refining your thought process through multiple iterations. Structure your response into two sections: Thought and Solution. In the Thought section, present your reasoning using the format: \texttt{"<think>\textbackslash n \{thoughts\} </think>\textbackslash n"}. Each thought should include detailed analysis, brainstorming, verification, and refinement of ideas. After \texttt{"</think>\textbackslash n,"} in the Solution section, provide the final, logical, and accurate answer, clearly derived from the exploration in the Thought section. If applicable, include the answer in \texttt{\textbackslash boxed\{\}} for closed-form results like multiple choices or mathematical solutions.

\textbf{User:} This is the problem:

\{Question\}

\textbf{Assistant:} \texttt{<think>}
\end{small}
\end{tcolorbox}
\caption{Chat template used for both Qwen2.5-Math-7B and Llama-3.1-8B across all experiments.}
\label{fig:chat-template}
\end{figure}

% \newpage
% \input{checklist.tex}

\end{document}